\documentclass[runningheads]{llncs}
\usepackage[T1]{fontenc}
\usepackage{graphicx}
\usepackage{cite}
\usepackage{hyperref}
\usepackage{nicefrac}
\usepackage{url}            
\usepackage{booktabs}       
\usepackage{amsfonts}       
\usepackage{nicefrac}       
\usepackage{microtype}      
\usepackage{xcolor}         
\usepackage{algorithmic}
\usepackage{algorithm}
\usepackage[misc,geometry]{ifsym}

\usepackage{subcaption}
\usepackage{threeparttable}
\usepackage{multirow}

\usepackage{amsmath}
\usepackage{mathtools}

\usepackage[capitalize,noabbrev]{cleveref}

\usepackage{pifont}

\DeclareMathOperator{\GNN}{GNN}

\begin{document}
\title{\textsc{Moco}: A Learnable Meta Optimizer for Combinatorial Optimization}
%
%
\author{Tim Dernedde \Letter \and
Daniela Thyssens \and
Sören Dittrich \and
Maximilian Stubbemann \and
Lars Schmidt-Thieme
}
\authorrunning{T. Dernedde et al.}
%
\institute{
    {
        Institute of Computer Science, University of Hildesheim, Hildesheim, Germany 
        \email{\{dernedde,thyssens,schmidt-thieme\}@ismll.uni-hildesheim.de}
        \email{\{dittrich,stubbemann\}@ismll.de}
    }
}
%
\maketitle              
\begin{abstract}
    Relevant combinatorial optimization problems (COPs) are often NP-hard. While
    they have been tackled mainly via handcrafted heuristics in the past, advances
    in neural networks have motivated the development of general methods to
    \emph{learn} heuristics from data. Many approaches utilize a neural network to
    directly construct a solution, but are limited in further improving based on already constructed solutions at inference time. Our approach, \textsc{Moco}, defines a lightweight solution construction procedure, guided by a single continuous vector $\theta$ (called heatmap) and learns a neural network to update $\theta$ for a single instance of a COP at inference time. The update is based on various features of the current search state. The training procedure is budget aware, targeting the overall best solution found during the entire search. \textsc{Moco} is a fully 
    learnable meta optimizer not utilizing problem specific heuristics or requiring optimal solutions for training. We test \textsc{Moco} on the Traveling Salesman Problem (TSP) and Maximum Independent Set (MIS) and show that it significantly improves over other heatmap based methods.\footnote{A prior version was published in Advances in Knowledge Discovery and Data Mining. PAKDD 2025. Lecture Notes in Computer Science, vol 15872. Springer, Singapore. }
    \keywords{COPs \and RL \and TSP \and MIS.}
\end{abstract}
\section{Introduction}%
\label{sec:introduction}%
Combinatorial optimization problems (COPs) underlying relevant applications
are often NP-hard. Thus, heuristics have been the primary
paradigm of tackling them. However, they are often handcrafted
and developed for
each COP individually, limiting the transferability of a heuristic from one COP to
another. This motivates the development of general methods which automatically
\emph{learn} heuristics. Driven by advances in neural networks, there has
been a resurgence of interest in data-driven heuristics for
combinatorial optimization. This emerging field is commonly referred to as
\emph{Neural Combinatorial Optimization} (NCO) \cite{Bello2017.Neural}.

Typically, NCO makes the assumption that there exists a distribution over
instances of a (class of) COP(s) for which a heuristic needs to be developed. One of the main paradigms of NCO solvers is to define a neural network guided solution construction process, that sequentially decides on values of decision variables until a complete solution is constructed. In the MIS problem for instance, at every step an additional node is added to the current independent set until no further nodes can be added.
Since constructing optimal solutions in one shot is difficult, additional search is needed. Here, further heuristic concepts, such as local search \cite{DBLP:conf/aaai/FuQZ21,Li2023.Distribution,Qiu2022.DIMES,Sun2023.DIFUSCOa,Ye2023.DeepACOa}, large neighborhood search \cite{DBLP:conf/ecai/HottungT20,Falkner2023.Too,Luo2023.Neural}, or tree search \cite{Bother2022.What,DBLP:conf/aaai/FuQZ21,DBLP:conf/nips/ChooKKJHTG22,Li2018.Combinatorial} are often used. However, such techniques are partially again problem specific and thus limit transferability. Additionally, when used in conjunction with neural networks, limited information about the search process is passed back to the network, making it difficult to update the solution construction process at inference time. 

We improve on these aspects with \textsc{Moco}, which uses a lightweight construction process, only guided by a single continuous vector $\theta$. This vector contains one value per decision variable and models the likelihood of certain variables to be included in the solution. It is sometimes referred to as a \emph{heatmap}.
Given a single instance, one needs to optimize over $\theta$ to construct optimal solutions for that instance. For this purpose we \emph{learn} a meta optimizer that takes information such as previously constructed solutions, information about the instance, the remaining optimization budget and more in order to update $\theta$. We follow the literature \cite{Metz2022.VeLO,Metz2019.Understanding} and use evolutionary strategies \cite{Salimans2017.Evolution} to train \textsc{Moco}, targeting the best overall solution found during the search.

This learned optimizer has additional favorable properties such as the
ability to optimize for a given inference budget as well as to adapt its strategy
to a range of different budgets not seen during training. Our contributions are:

\begin{enumerate}%
    \item We propose \textsc{Moco}, a \textbf{novel learnable optimizer} for combinatorial optimization that can learn to update an instance specific construction process based on feedback from previously constructed solutions.
    \item Our model is trained with an \textbf{inference budget aware} meta learning procedure. We show that this training procedure is effective such that higher budgets lead to better solutions. Additionally, we show that \textsc{Moco} learns to \textbf{generalize to different budgets} not seen during training and conditioning on higher budgets than seen during training also improves solution quality. Such budget conditioning is a novel feature in the NCO literature.
    \item Our experiments further show that learning to update the construction procedure improves significantly over other methods that utilize the same construction procedure. Additionally, we outperform methods on final solution quality, that do not rely on optimal solutions for supervised training. Code and checkpoints are available at \url{https://github.com/TimD3/Moco}.
\end{enumerate}%

\begin{figure}%
    \centering
    \includegraphics[width=\textwidth]{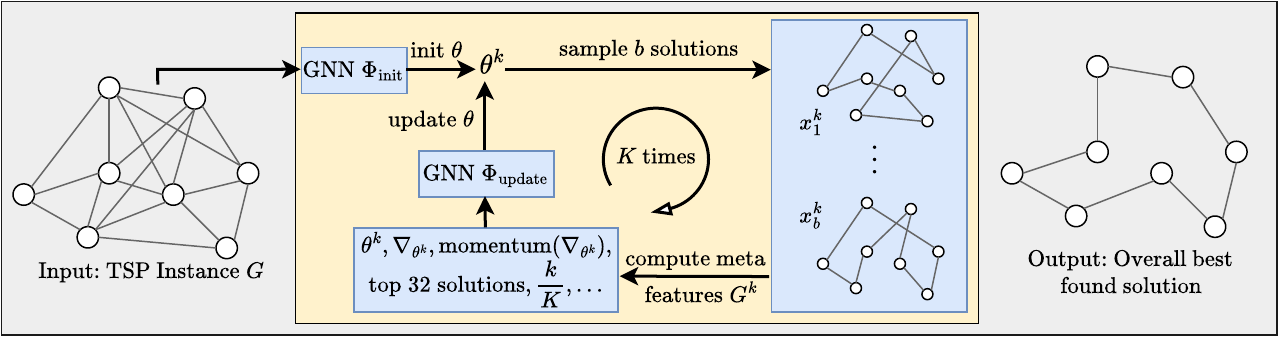}
    \caption{Overview of our approach with the TSP as an example. Given an instance as a graph, $\GNN_{\Phi_{\text{init}}}$ initializes $\theta$. Afterwards the following process is repeated $K$ times: First, $b$ solutions are sampled from $\theta$. Second, meta features are constructed based on the solutions and previous iterations and third, $\GNN_{\Phi_{\text{update}}}$ updates $\theta$.}
    \label{fig:overview1}
\end{figure}%

\section{Related Work}%
\label{sec:related_work}%
\subsubsection{Construction from neural networks}%
Construction from neural networks refers to methods that directly construct solutions
for discrete optimization problems by employing a neural network which sequentially makes decisions on the value of a decision variable (or multiple variables) until a complete solution is
constructed. The neural network computes a probability distribution over the next decision variable(s) to be set. Such approaches have been applied to a variety of problems such as the TSP \cite{Jin2023.Pointerformer, DBLP:conf/cp/JoshiCRL21, DBLP:conf/iclr/KoolHW19, DBLP:conf/nips/KimPP22, Kwon2020.POMO},
different routing problems \cite{DBLP:conf/iclr/KoolHW19, Kwon2020.POMO}, scheduling problems \cite{Grinsztajn2023.Winner},
and other traditional problems such as Knapsack \cite{Drakulic2023.BQNCO, Ye2023.DeepACOa} and MIS \cite{Ahn2020.Learninga}.

They often work well on small problems, however since the neural network is computed at every construction step, they are computationally expensive and difficult to train to good
solution quality with reinforcement learning for larger problem sizes \cite{DBLP:conf/cp/JoshiCRL21}. Thus, the
literature has partially focused on improving and evaluating size extrapolation,
where a model gets trained on a small problem size and then evaluated on a
larger problem size \cite{DBLP:conf/cp/JoshiCRL21, DBLP:conf/iclr/HottungKT22, DBLP:conf/nips/ChooKKJHTG22, Falkner2023.Too, Grinsztajn2023.Winner, Luo2023.Neural}. 
\subsubsection{Construction from heatmaps}
Heatmap approaches are methods where a neural network predicts a vector of values, called heatmap, for all decision variables in one shot. Each entry is assigned to one of the decision variables of the COP. The heatmap can then be used to guide solution construction or further search procedures.

One line of research models each decision variable independently, predicting the probability of it being set to $1$, i.e. belonging to the optimal solution \cite{DBLP:journals/corr/abs-1906-01227, DBLP:conf/aaai/FuQZ21, Sun2023.DIFUSCOa, Li2023.Distribution, Li2018.Combinatorial, Bother2022.What}. These methods are learned in a supervised way from optimal solutions, either as a classification problem \cite{DBLP:journals/corr/abs-1906-01227, DBLP:conf/aaai/FuQZ21} or via diffusion modelling \cite{Sun2023.DIFUSCOa, Li2023.Distribution}.

Another line of work models the heatmap as an unnormalized probability distribution over all decision variables and constructs solutions by autoregressive sampling \cite{Qiu2022.DIMES, Ye2023.DeepACOa}. Training was done via reinforcement learning.

At inference time, heatmaps can be further updated towards higher quality solution regions via policy gradients \cite{Qiu2022.DIMES}, ant colony strategies \cite{Ye2023.DeepACOa, Dorigo2006.Ant}, MCTS \cite{DBLP:conf/aaai/FuQZ21} and further surrogate objectives \cite{Li2023.Distribution}. Additionally, most methods can optionally be hybridized with further search procedures such as local search or dynamic programming \cite{Kool2022.Deep, Dorigo2006.Ant, DBLP:conf/aaai/FuQZ21, Sun2023.DIFUSCOa, Li2023.Distribution, Qiu2022.DIMES, Ye2023.DeepACOa}. 
We summarize this categorization of heatmap based approaches in \cref{tab:categorization}.

\begin{table}[t]
    \caption{Categorization of Heatmap based approaches: We classify the methods based on how they initialize, update and decode the heatmaps, as well as the utilized training procedures. If both initialization and updating of the heatmap are done in a heuristic fashion, one yields an Ant-Colony optimization approach, giving a classical counterpart to heatmap based approaches. [$\cdot$] refers to optional components.}
    \label{tab:categorization}
    \centering
    \resizebox{\textwidth}{!}{%
        \begin{tabular}{l|llll}
            Method                                        & Initialization & Update              & Decoding                      & Training                \\
            \hline
            Ant Colony Optimization \cite{Dorigo2006.Ant} & Heuristic      & Heuristic           & Sampling, [LS]                & -                       \\
            \hline
            GCN \cite{DBLP:journals/corr/abs-1906-01227}  & Learned        & -                   & Beam Search                   & Supervised              \\
            Att-GCN \cite{DBLP:conf/aaai/FuQZ21}          & Learned        & MCTS                & k-opt                         & Supervised              \\
            Difusco \cite{Sun2023.DIFUSCOa}               & Learned        & -                   & Greedy, [Multistart, 2-opt]   & Supervised, Diffusion   \\
            T2T \cite{Li2023.Distribution}                & Learned        & Surrogate Objective & Greedy, [Multistart, 2-opt]   & Supervised, Diffusion   \\
            \hline
            Dimes \cite{Qiu2022.DIMES}                    & Learned        & REINFORCE + Adam    & Sampling, [k-opt]             & REINFORCE + Maml        \\
            DeepACO \cite{Ye2023.DeepACOa}                & Learned        & Heuristic           & Sampling, [2-opt]             & REINFORCE               \\
            \textsc{Moco} (Ours)                          & Learned        & Learned             & Sampling, [Multistart, 2-opt] & Evolutionary Strategies \\
        \end{tabular}
    }
\end{table}
\subsubsection{Learned Optimizers and Meta Learning for CO}
Meta Learning optimizers for continuous tasks such as training neural networks and displacing handcrafted optimizers such as Adam 
 has long been a topic of interest \cite{Bengio1992.Optimization, Andrychowicz2016.Learning}. We draw on recent work \cite{Metz2019.Understanding, Metz2022.VeLO} and connect our method to this field in two ways: First, we formulate discrete graph combinatorial problems as a continuous optimization problem by defining a solution construction process, controlled by a continuous vector $\theta$ which can be updated via the REINFORCE gradient at inference time. This allows the design of a learned meta optimizer that can take into account additional problem specific structures of interest to us, in contrast to mentioned work, which is interested in learning optimizers mainly for neural networks \cite{Metz2022.VeLO}. Second, we adopt their training methods based on evolutionary strategies \cite{Salimans2017.Evolution}.

Note that the term \emph{meta learning} can have different meanings. We consider our system to be meta learned since there are two layers of optimization. This is in line with the use of the term in the \emph{learning to optimize} field, as discussed by \cite{JMLR:v23:21-0308}. 

Our method is focused on learning a model to solve instances from a single problem class and distribution. In contrast, other work in the NCO community uses methodologies from meta learning with the differing goal of achieving quick adaption to new distributions of instances or objectives \cite{Chen2023.Efficient, DBLP:conf/icml/0002WSCZ23, DBLP:conf/icml/SonKKP23}.

\section{Methodology}
\label{sec:methodology}
\subsubsection{Problem Statement}
\label{sec:problem_statement}
Our work considers NP-hard binary combinatorial optimization problems (COPs) formulated over graphs. Let $G=(V,E)$ be a graph, where $V$ are the vertices and  $E \subseteq V \times V$ are the edges of $G$. A COP consists of a set $\Omega_G \subseteq \{0,1\}^N$ of \emph{feasible solutions} and an \emph{objective function} $f_G \colon \Omega_G \to \mathbb{R}$. The goal is to find an \emph{optimal feasible solution} $x=(x_1,\dots, x_N) \in \Omega_G$, minimizing $f_G$: $\min_{x \in \Omega_G} f_G(x).$
The elements  $x_i \in \{0,1\}$ are also referred to as the \emph{binary decision variables}.
Typically, the problem involves the selection of either edges or nodes in the graph leading to either $N=|E|$ or $N=|V|$. For example, two prominent COPs are the Traveling Salesman Problem (TSP) and the Maximum Independent Set (MIS) problem. In the TSP, a subset of edges is selected such that the Hamiltonian cycle with minimum distance is found, leading to one decision variable per edge. In MIS, we need to find the set of independent nodes of maximal size, resulting in one decision variable per node. An extended description of both problems can be found in \cref{app:tsp} and \ref{app:mis}.
\subsubsection{Construction Process}
\label{sec:construction_process}
We solve such problems with a constructive process that sequentially decides on the values of the decision variables. Given a \textbf{single}
instance of a COP with $N$ binary decision variables, we instantiate a shallow
policy $\pi_\theta$. To be more specific, our policy only depends on a
so-called \emph{heatmap} $\theta \in \mathbb{R}^{N}$ with one parameter per decision variable. The policy constructs solutions from $\theta$ as follows.

The process starts with $x^0=0$. Subsequently, at each construction step $t$, the
policy emits a probability distribution $p_\theta(a^t \mid x^ {t})$, where $a^t
    \in \{1, \dots, N\}$ represents the next decision variable to be set to $1$
given the current partial assignment $x^{t}$. Here, $p_\theta$ is computed by
normalizing $\theta$ via softmax. Additionally,
actions that given $x^{t}$ would lead to infeasible solutions or have already
been set to 1 before, are masked away. In detail, the probability that $a_t=i$ is given by
\begin{equation*}
    p_\theta(a^t=i \mid x^{t}) = \frac{\exp(\mathcal{M}(\theta)_i)}{\sum^N_{j=1} \exp(\mathcal{M}(\theta)_j)}, \mathcal{M}(\theta)_i = \begin{cases}
        -\infty  & \text{if } x^{t}_i=1 ,                                 \\
        -\infty  & \text{if } x^{t+1}_i=1  \Rightarrow \text{ infeasible}, \\
        \theta_i & \text{otherwise}.
    \end{cases}
\end{equation*}
Here, "$x_i^{t+1}=1 \Rightarrow \text{ infeasible}$" refers to the situation, where it is
not possible to proceed to a feasible solution $x \in \Omega_G$ from the current
state.
The policy then samples from $p_\theta(a^t \mid x^{t})$ and sets the corresponding
decision variable to $1$ leading to $x^{t+1}$, which is repeated until
termination at step $T$. We repeat this process $b$ times and arrive
at $b$ solutions.
This allows solving the COP by maximizing the negative expected objective $J(\theta) = \mathbb{E} [-f(x^{T})]$ with policy gradients.
Using REINFORCE 
 with an average reward baseline
leads to
\begin{equation}\label{eq:policy_gradient}
    \nabla_\theta J(\theta) = -\mathbb{E} \left[ \sum_{t=1}^T (f(x^{T})- \mathbb{E} [ f(x^{T}) ]) \nabla_{\theta} \log p_{\theta}(a^t|x^t) \right].
\end{equation}

In the case of the TSP we make an enhancement to the generic construction process
by limiting the next variables to be decided over only to the
edges connected to the current node. Here, the current node is determined by the
last edge that was chosen, and the
starting node is chosen randomly. This limits the number of variables the
softmax is computed over and is common in the literature
\cite{Jin2023.Pointerformer, Drakulic2023.BQNCO, Qiu2022.DIMES, Ye2023.DeepACOa}. Note, that our
general approach would also work without this problem-specific modification
which is only done to restrict the search space.
\subsubsection{Meta Optimizer}\label{sec:meta_optimizer}
Compared to constructing solutions with a policy parametrized by a deep neural network, $\pi_\theta$ is limited in expressivity. However, its simplicity allows computing solutions more efficiently than a deep neural network.

In principle, any gradient based optimizer can be used to optimize $\theta$. In \cite{Qiu2022.DIMES} 
 for instance, $\theta$ is initialized via a GNN and subsequently updated with Adam
. This approach however has multiple potential shortcomings.
First, it mostly ignores the problem specific structure, meaning there is little guidance on how good solutions for this problem class typically look like, which solutions have already been explored and which areas of the solution are still worth exploring.
Second, there is little consideration for the computational budget. Since this is a search procedure happening at inference time, there is a computational budget set. The search procedure should take the time budget into account, which in this procedure can only be done in a limited fashion.

To tackle these shortcomings, we propose a learnable meta optimizer that initializes and updates $\theta$ after every batch of $b$ constructed solutions.
Our learnable optimizer consists of two graph neural networks $\GNN_{\Phi_{\text{init}}}$ and $\GNN_{\Phi_{\text{update}}}$ which are parametrized by two sets of parameters $\Phi = \{\Phi_{\text{init}}, \Phi_{\text{update}}\}$. Given a COP consisting of $G,\Omega_G,f_G$, we use $\GNN_{\Phi_{\text{init}}}$ to initialize $\theta^{0}$.
Then, after constructing a batch of solutions from $\pi_{\theta^0}$, the second network $\GNN_{\Phi_{\text{update}}}$ updates $\theta^0$ based on the solution and additional features. This process is repeated $K$ times. In detail, at each iteration $k$, we concatenate the following features to the graph $G$:
\begin{enumerate}
    \item The values $\theta^{k}$, $\nabla_{\theta^{k}}$ based on Eq. \ref{eq:policy_gradient} and 6 momentum values of $\nabla_\theta^k$ with $\beta = 0.1, 0.5, 0.9, 0.99, 0.999, 0.9999$, which gives the learned optimizer information about the current parameters and several estimates of good descent directions.
    \item The top $L=32$ solutions $x_1, ..., x_L$ found so far and the corresponding objective function values $f_G(x_l)$ normalized via $\nicefrac{(f_G(x_l)-f_G(x_1))}{f_G(x_1)}$.
    \item The relative improvement over the previous step $\nicefrac{(f_G(x^{k-1}_\text{best})-f_G(x^{k}_\text{best}))}{f_G(x^{k}_\text{best})}$.
    \item The current iteration $k$ encoded with a tanh-embedding $\tanh({\nicefrac{k}{\text{timescale}}-1})$, where $\text{timescale} = 1, 3, 10, 30, 100, 300, 1000, 3000, 10000, 30000, 100000$, following \cite{Metz2022.VeLO}. 
    Additionally, we use the relative fraction $\frac{k}{K}$ as a feature.
\end{enumerate}
Note that all these features represent node features, edge features or global graph features.
Thus, we can employ $GNN_{\Phi_\text{update}}$ to parametrize the update-rule.
We denote the input to $GNN_{\Phi_\text{update}}$ at iteration $k$ as the feature graph $G^{k}$.

The GNN outputs a global graph representation and individual node and/or edge
representations, depending on the COP, which are decoded to
$\tilde{\theta}^{k}$ by a linear layer. Additionally, the global graph features are processed through another linear layer to
derive a scalar $\alpha^{k} \in \mathbb{R}$. Then the update rule for $\theta$ is given by 
\begin{equation}\label{eq:theta_update}
    \theta^{k+1} = \frac{\tilde{\theta}^{k}}{\alpha^{k}}
\end{equation}
The scalar $\alpha$ helps to more easily control the entropy of the distribution induced by $\theta^{k+1}$. Our GNN thus emits the new parameters in a direct fashion, which is in contrast
to traditional gradient based optimizers and learnable optimizers for neural
networks \cite{Metz2022.VeLO,Metz2019.Understanding}, which use
additive update rules. In preliminary experiments we have found this to be
slightly better
performing. We hypothesize this to be related to the simplicity of $\theta$
and the ability to more easily control a budget dependent temperature scaling strategy through $\alpha$.

\begin{algorithm}[tb]
    \caption{\textsc{Moco} Inference}\label{alg:two}
    \begin{algorithmic}
        \STATE {\bfseries Input:} Graph $G$, Trained GNNs $\GNN_{\Phi_{\text{init}}}$
        and $\GNN_{\Phi_{\text{update}}}$, Budget $K$, Batch Size $b$
        \STATE $\theta^{(0)} \gets \GNN_{\Phi_{\text{init}}}(G)$
        \COMMENT{\textit{initialize} $\theta$ \textit{via GNN}}
        \FOR{$k\gets0$ to $K-1$}
        \STATE $\text{solutions} \gets \text{rollout}(\theta^{k}, b)$ \COMMENT{\textit{sample a batch of solutions}}
        \STATE $x^{k}_{\text{best}} \gets \text{update\_best\_solution}(\text{solutions})$\;
        \STATE $G^{k} \gets \text{construct\_feature\_graph}(\theta^{k}, k, K, \text{solutions})$ \COMMENT{\textit{compute input features for the meta optimizer as described in section \ref{sec:meta_optimizer}}}
        \STATE $\Tilde{\theta^{k}}, a^{k} \gets \GNN_{\Phi_{\text{update}}}(G^{k})$\;
        \STATE $\theta^{k+1} = \frac{\Tilde{\theta}^{k}}{a^{k}}$ \COMMENT{\textit{update} $\theta$ \textit{via \cref{eq:theta_update}}}
        \ENDFOR

        \STATE {\bfseries Return:} $x^{K-1}_{\text{best}} $
    \end{algorithmic}
\end{algorithm}

\subsubsection*{Model Architecture}
We utilize the same architecture for $\GNN_{\Phi_{\text{init}}}$ and $\GNN_{\Phi_{\text{update}}}$.
For the TSP we build our architecture on the \emph{GraphNetwork} framework in \cite{Battaglia2018.Relational}, computing node and edge embeddings. 
 For MIS, we use a GCN~\cite{Kipf2017.Semisupervised} with additional global graph feature updates.
 For a detailed description of the models used, see  \cref{app:architecture}. \cref{alg:two} and 
\cref{fig:overview1} summarizes the overall described procedure.

\subsubsection{Meta Training}\label{sec:meta_training}
We train our model over different instances of
the COP defined by different graphs. We thus assume that our graphs are sampled
from a distribution $D$: $G\sim D$. Our meta objective can be specified via
\begin{equation*}
    L(\Phi) = \mathbb{E}_{G \sim D} \left[ f_G(x^{K-1}_{\text{best}})\right],
\end{equation*}
where $x^{K-1}_{\text{best}}$ is the best found solution after $K$ steps (see \cref{alg:two}).
Since this objective is non-differentiable and results from a potentially long
optimization trajectory, we follow the related work on training meta optimizers
and employ evolutionary strategies (ES) \cite{Salimans2017.Evolution, Metz2022.VeLO} to optimize $\Phi$. ES is a black box
optimization method where the parameters are randomly perturbed multiple times
and then evaluated for their resulting objective value. A descent direction is
estimated as a weighted average of the perturbations scaled by their resulting
objective value. ES has been explored as a simple alternative to Reinforcement
Learning Algorithms and has been shown to be effective for training learned
optimizers even over exact gradient computation for two reasons
\cite{Metz2019.Understanding, Metz2022.VeLO}. Firstly, it is more memory efficient
since it does not require backpropagation through the optimization trajectory.
Secondly, the ES estimator (\cref{eq:es_estimator}) optimizes the Gaussian
smoothed objective $\mathcal{L}(\Phi) = \mathbb{E}_{\epsilon \sim \mathcal{N}(0,
    I)} \left[L(\Phi + \sigma^2 \epsilon)\right]$, where $I$ is the identity matrix. This  helps with possibly
unsuitable loss surfaces and the randomness in the stochastic meta objective
$L$, which has been observed in the past~\cite{Metz2019.Understanding}. We also
use antithetic sampling  leading to the
following estimator:
\begin{equation}\label{eq:es_estimator}
    \nabla_{\Phi} \mathcal{L}(\Phi) \approx \frac{1}{2\sigma^2N} \sum_{i=1}^{N/2} \sigma^2 \epsilon_i \left( L(\Phi + \sigma^2 \epsilon_i) - L(\Phi - \sigma^2 \epsilon_i) \right).
\end{equation}
Here, $N$ refers to the number of perturbations. Since each perturbation
$\epsilon_i \sim \mathcal{N}(0, I)$ gets used twice in \cref{eq:es_estimator},
we only sample $\frac{N}{2}$ different $\epsilon_i$. The hyperparameter $\sigma^2$ determines the strength
of the perturbation. We use $\sigma^2 = 0.01$.
\section{Experiments}\label{sec:experiments}
\subsection{Datasets, Metrics, Hardware and Baselines}
To evaluate our approach we test it on the TSP as an edge selection task and the MIS problem as a node selection task. For the TSP, we use the test datasets introduced in
\cite{DBLP:conf/aaai/FuQZ21}. 
 Nodes are sampled uniformly at random from the
2-dimensional unit square and instances consist of 100,
200 and 500 nodes. Pairwise distances are determined via euclidean distance. For training, instances are generated on the fly. In line with the related work
\cite{DBLP:conf/aaai/FuQZ21, Qiu2022.DIMES, Ye2023.DeepACOa}, the graphs are sparsified to the 20 nearest neighbors for TSP100, TSP200 and the 50 nearest neighbors for TSP500 in order to reduce computational cost.
We report the solution cost and relative gap in percent to the reference solution from LKH-3 \cite{lkh3} averaged over the dataset.
For MIS, we follow the experimental protocol in \cite{Qiu2022.DIMES} 
and consider three datasets. For two, instances are generated from the Erd\H{o}s-R\'{e}nyi (ER) random graph model $G(n, p)$, where $n$ is the number of nodes and $p$ is the probability of an edge between two nodes. We generate instances in two distributions. The smaller contains graphs with sizes uniformly at random between 700 and 800 nodes and $p=0.15$. The larger contains graphs with 9000 to 11000 nodes and $p=0.02$. For training, we generate 4096 instances of ER-[700-800]. The ER-[9000-11000], graphs are too large to effectively train our approach with our hardware budget, so we rely on the generalization behavior from the smaller graphs. The last dataset contains 40000 graphs reduced from SATLIB instances. The test datasets for the ER graphs are directly taken from \cite{Qiu2022.DIMES} 
 and for SATLIB, we follow the specified data split.
We report the independent set size and relative gap in percent to the reference solution by KaMIS \cite{Lamm2017.Finding}  averaged over the dataset.

Inference is done on a node with a single RTX 3090 GPU and EPYC 7543P CPU
. For \cref*{tab:mis_main,tab:tsp_main,tab:hybrid}, we also report the average runtime in seconds to solve a test set instance. If not otherwise stated, we rerun the baseline methods from the publicly available code and checkpoints on our hardware.

We test against a variety of baselines, divided into: (i) traditional OR methods, (ii) ML-based methods using supervised learning, which require a solver to generate training data with (near)-optimal solutions, and (iii) ML-based methods that do not rely on supervised learning.
For more details on the individual baselines, see \cref{app:baselines}.
\subsection{Generalization over $b$, $K$ and comparison to heatmap methods}
\begin{figure}[t]%
    \begin{subfigure}{0.5\columnwidth}
        \centering
        \includegraphics[height=0.21\textheight]{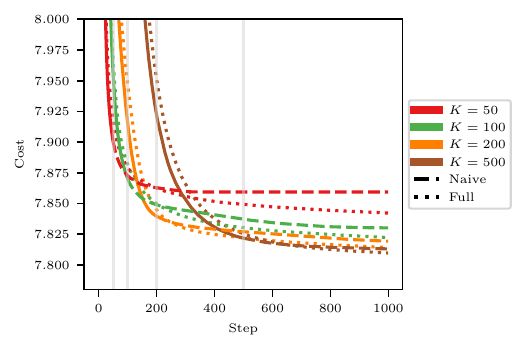} 
        \captionsetup{width=0.97\textwidth}
        \caption{Extrapolation behavior to budget $K$.}
        \label{fig:behaviour_extrapolation}
    \end{subfigure}\hfill
    \begin{subfigure}{0.5\columnwidth}
        \centering
        \includegraphics[height=0.21\textheight]{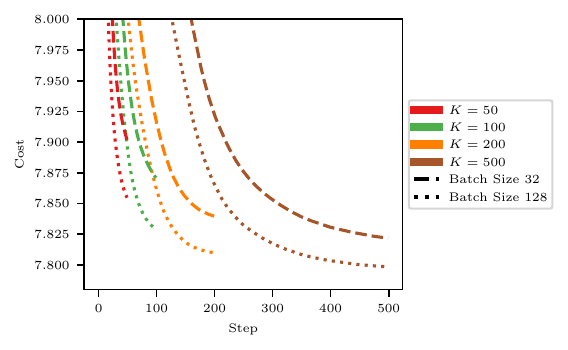} 
        \captionsetup{width=0.97\textwidth}
        \caption{Extrapolation behavior to batch size $b$.}
        \label{fig:behaviour_batch_size}
    \end{subfigure}
    \caption{Learned optimizer behavior on the TSP100. All
        optimizers are conditioned by a relative budget feature $\nicefrac{k}{K}$. (a) shows optimizers trained on differing budgets of $K^{\text{train}}=50,100,200,500$ and executed for $K^{\text{inference}}=1000$. During inference, the relative budget feature is then either calculated as (i) $\nicefrac{k}{K^{\text{train}}}$ (Naive), leading to values outside the range seen during meta training or (ii) as $\nicefrac{k}{K^{\text{inference}}}$ (Full). (b) shows the behavior of optimizers when tested on larger batch sizes than seen during training, where $b=32$ was used during training.
        Note that these experiments were done with a preliminary version of our model that lacked $\GNN_{\Phi_{\text{init}}}$.
    }
    \label{fig:behaviour}
\end{figure}

Figure \ref*{fig:behaviour} displays the behavior learned by our meta optimizer.
In \cref{fig:behaviour_extrapolation} we can see that when trained for different
budgets of $K$, the optimizer directly trained for this budget also demonstrates
the best performance over the others after exactly $K$ steps. This shows that the
meta training procedure is generally effective and \textsc{Moco} can learn budget dependent strategies
for trading off
exploration and exploitation. Additionally, all learned optimizers
show adaptive behavior, where when conditioned on a larger budget at inference time than used during training, they will further improve solution quality.  However, the best performance is still achieved when training and targeted inference budget match. \cref{fig:behaviour_batch_size} further demonstrates, that the
optimizers are also robust against an increase in the batch size, making
effective use of larger batch sizes than seen during training.

\begin{table}%
    \centering
    \caption[short]{Detailed comparison of \textsc{Moco} to other heatmap methods on the TSP500 with different budgets $K$ and number of samples per step $b$. All methods were tested with one run per instance $M=1$ and additional local search deactivated.
    }\label{tab:comparison_dimes}
    \resizebox{0.7\textwidth}{!}{%
        \begin{tabular}{ll|llllllll}
            \multirow{ 2}{*}{$K$}  & \multirow{ 2}{*}{$b$} & \multicolumn{2}{c}{Aco} & \multicolumn{2}{c}{DeepACO} & \multicolumn{2}{c}{Dimes} & \multicolumn{2}{c}{\textsc{Moco}}                                                                               \\
                                   &                       & Cost $\downarrow$       & Gap $\downarrow$            & Cost $\downarrow$         & Gap $\downarrow$                  & Cost $\downarrow$ & Gap $\downarrow$ & Cost $\downarrow$ & Gap $\downarrow$ \\
            \hline
            \multirow{ 5}{*}{200 } & 32                    & 23.95                   & 44.65\%                     & 18.71                     & 13.01\%                           & 18.06             & 9.09\%           & \textbf{17.23}             & \textbf{4.07\%}           \\
                                   & 64                    & 23.64                   & 42.81\%                     & 18.63                     & 12.56\%                           & 17.96             & 8.51\%           & \textbf{17.10}             & \textbf{3.32\%}           \\
                                   & 128                   & 23.49                   & 41.87\%                     & 18.58                     & 12.22\%                           & 17.89             & 8.06\%           & \textbf{17.06}             & \textbf{3.03\%}           \\
                                   & 256                   & 23.33                   & 40.90\%                     & 18.52                     & 11.87\%                           & 17.80             & 7.55\%           & \textbf{17.00}             & \textbf{2.67\%}           \\
                                   & 512                   & 23.16                   & 39.88\%                     & 18.48                     & 11.65\%                           & 17.75             & 7.19\%           & \textbf{16.96}             & \textbf{2.46\%}           \\
            \hline
            \multirow{ 5}{*}{500 } & 32                    & 23.43                   & 41.51\%                     & 18.61                     & 12.41\%                           & 18.02             & 8.87\%           & \textbf{17.19}             & \textbf{3.83\%}           \\
                                   & 64                    & 23.19                   & 40.05\%                     & 18.55                     & 12.08\%                           & 17.95             & 8.45\%           & \textbf{17.08}             & \textbf{3.16\%}           \\
                                   & 128                   & 23.02                   & 39.04\%                     & 18.49                     & 11.69\%                           & 17.88             & 7.98\%           & \textbf{17.03}             & \textbf{2.88\%}           \\
                                   & 256                   & 22.90                   & 38.31\%                     & 18.45                     & 11.43\%                           & 17.79             & 7.46\%           & \textbf{16.98}             & \textbf{2.55\%}           \\
                                   & 512                   & 22.74                   & 37.37\%                     & 18.39                     & 11.11\%                           & 17.72             & 7.07\%           & \textbf{16.93}             & \textbf{2.25\%}           \\
        \end{tabular}%
    }
\end{table}
\subsection{TSP and MIS Results}
Results for our method are reported with varying numbers of steps $K$ and budgets $b$. Additionally, we introduce a parameter $M$, denoting the number of repeated optimization runs for a single instance. This is similar to the multiple restarts used by other methods. Note that all of them are always done in parallel on the GPU together. For the TSP, we train our model with $K=200$ and $b=32$ for all sizes. For MIS, we train two models for the small ER graphs with $K=50, 200$ and $b=32$. For SATLIB, we only train one model with $K=50$ and $b=32$.

Results on the TSP are shown in Table \ref*{tab:tsp_main}. \textsc{Moco} outperforms related methods that also construct from heatmaps, such as Dimes, DeepACO, Difusco and T2T on solution quality. Among methods that construct with neural networks we also outperform the Pointerformer and Pomo with and without additional Tree and Active Search in terms of solution quality. Only the BQ-Transformer achieves better solution quality. Its training however relies on supervised learning from LKH solutions which strongly limits the applicability to other problems.

Table \ref*{tab:mis_main} shows the results of our MIS experiments. \textsc{Moco} significantly outperforms other learning-based approaches on the smaller ER graphs, reducing the optimality gap to $0.85\%$. On the ER-[9000-11000], we cannot train \textsc{Moco} with our hardware budget given the graph sizes. Thus, we rely on the generalization performance of the model trained on ER-[700-800]. Despite the distributional shift, due to the large graphs being an order of magnitude larger and having a reduced edge connection probability (from $p=0.15$ to $p=0.02$), we still achieve a low gap of 4.64\%. For the SATLIB dataset, all methods are close together, with our method achieving the lowest gap among the non supervised models. Results could be further improved by finetuning the model on $K=200$.

A limitation of \textsc{Moco} on the large-scale MIS graphs is the runtime. For problems with many decision variables and construction steps, the computation of the policy gradient dominates, and the procedure becomes slow. However, this can be mitigated by not constructing entire solutions at every step, but similar to LNS strategies, fixing a subset of decision variables based on the current best solution and only optimizing $\theta$ over the remaining ones. This would reduce both the number of variables the gradient is computed over and the number of steps needed to construct a solution. Additionally, such localized optimization might even improve solution quality further. We leave this for future work.
\begin{table}
    \centering
    \caption{TSP results without hybrid heuristics.}\label{tab:tsp_main}
    \resizebox{\textwidth}{!}{
        \begin{tabular}{lll|lllllllll} 
                                                                & \multirow{ 2}{*}{Method }                         & \multirow{ 2}{*}{Dec. Str.}                                               & \multicolumn{3}{c}{100} & \multicolumn{3}{c}{200} & \multicolumn{3}{c}{500}                                                                                                                       \\
                                                                &                                                   &                                                                           & Cost $\downarrow$       & Gap $\downarrow$        & Time $\downarrow$       & Cost $\downarrow$ & Gap $\downarrow$ & Time $\downarrow$ & Cost $\downarrow$ & Gap $\downarrow$ & Time $\downarrow$ \\
            \hline
            \multirow{2}{*}{\rotatebox[origin=c]{90}{OR}}       & LKH3  \cite{lkh3}                                 &                                                                           & 7.7609                  & 0.000\%                 & 5.19                    & 10.719            & 0.00\%           & 6.37              & 16.555            & 0.00\%           & 10.28             \\
                                                                & Farthest Insertion                                &                                                                           & 9.9200                  & 27.819\%                & N/A                     & 13.954            & 30.17\%          & N/A               & 21.753            & 31.40\%          & N/A               \\
            \hline
            \multirow{3}{*}{\rotatebox[origin=c]{90}{Sup.}}     & Difusco   \cite{Sun2023.DIFUSCOa}                 & S                                                                         & 7.7778                  & 0.217\%                 & 1.56                    & N/A               & N/A              & N/A               & 17.670            & 6.74\%           & 7.92              \\
                                                                & T2T  \cite{Li2023.Distribution}                   & S                                                                         & 7.7653                  & 0.056\%                 & 1.83                    & N/A               & N/A              & N/A               & 17.051            & 3.00\%           & 4.53              \\
                                                                & BQ  \cite{Drakulic2023.BQNCO}                     & BS64                                                                      & \textbf{7.7610}                  & \textbf{0.001\%}                 & 0.96                    & \textbf{10.723}            & \textbf{0.04\%}           & 2.51             & \textbf{16.618}            & \textbf{0.38\%}           & 26.86             \\
            \hline
            \multirow{9}{*}{\rotatebox[origin=c]{90}{Not Sup.}} & DeepAco \cite{Ye2023.DeepACOa}                    &                                                                           & 8.1854                  & 5.469\%                 & 7.96                    & N/A               & N/A              & N/A               & 18.39             & 11.11\%          & 112.45            \\
                                                                & Dimes \cite{Qiu2022.DIMES}                        & AS + S                                                                    & 8.5484                  & 10.146\%                & 27.50                   & N/A               & N/A              & N/A               & 17.725            & 7.07\%           & 100.14            \\
                                                                & Pomo   \cite{Kwon2020.POMO}          & 8x aug + Pomo                                                             & 7.7730                  & 0.155\%                 & 0.06                    & 10.895            & 1.64\%           & 0.14              & 19.527            & 17.95\%          & 0.47              \\
                                                                & Pomo    \cite{Kwon2020.POMO}         & SGBS + EAS \cite{DBLP:conf/iclr/HottungKT22, DBLP:conf/nips/ChooKKJHTG22} & 7.7650                  & 0.053\%                 & 16.05                   & 10.782            & 0.58\%           & 71.52             & 18.461            & 11.52\%          & 379.17            \\
                                                                & Pointerformer  \cite{Jin2023.Pointerformer} & 8x aug + Pomo                                                             & 7.7732                  & 0.158\%                 & 0.09                    & 10.793            & 0.69\%           & 0.17              & 17.097            & 3.28\%           & 0.43              \\
                                                                & \textsc{Moco}                                     & M=1 K=200 b=32                                                            & 7.8409                  & 1.030\%                 & 1.07                    & 10.963            & 2.27\%           & 2.11              & 17.205            & 3.93\%           & 6.01              \\
                                                                & \textsc{Moco}                                     & M=1 K=200 b=128                                                           & 7.8100                  & 0.632\%                 & 1.08                    & 10.879            & 1.49\%           & 2.22              & 17.052            & 3.01\%           & 6.60              \\
                                                                & \textsc{Moco}                                     & M=32 K=200 b=32                                                           & 7.7667                  & 0.074\%                 & 2.61                    & 10.790            & 0.66\%           & 5.75              & 16.975            & 2.54\%           & 21.61             \\
                                                                & \textsc{Moco}                                     & M=32 K=500 b=128                                                          & \textbf{7.7622}                  & \textbf{0.016\%}                 & 6.24                    & \textbf{10.745}            & \textbf{0.24\%}           & 25.76             & \textbf{16.845}            & \textbf{1.75\%}           & 171.23            \\
        \end{tabular}%
    }
\end{table}
\begin{table}
    \caption{MIS Results:
        We follow \cite{Qiu2022.DIMES,Sun2023.DIFUSCOa} 
         and report results without the use of problem specific LS or Graph reduction techniques for ML baselines, since \cite{Bother2022.What} 
         reported that these techniques lead to near optimal solutions even from random solutions.
        * denotes results reported from \cite{Qiu2022.DIMES} ran on a single NVIDIA Ampere A100 40 GB. ** denotes results reported from \cite{Zhang2023.Let}, where for ER graphs we only report the relative gaps, since they use different test sets.
    }
    \label{tab:mis_main}
    \begin{threeparttable}
        \resizebox{\textwidth}{!}{%
            \begin{tabular}{lll|lllllllll}
                                                                     & \multirow{ 2}{*}{Method }          & \multirow{ 2}{*}{Dec. Str. } & \multicolumn{3}{c}{SATLIB} & \multicolumn{3}{c}{ER-[700-800]} & \multicolumn{3}{c}{ER-[9000-11000]}                                                                                                                           \\
                                                                     &                                    &                              & Set Size $\uparrow$        & Gap $\downarrow$                 & Time $\downarrow$                   & Set Size $\uparrow$ & Gap $\downarrow$ & Time $\downarrow$ & Set Size $\uparrow$ & Gap $\downarrow$ & Time $\downarrow$ \\
                \hline
                \multirow{ 2}{*}{\rotatebox[origin=c]{90}{OR}}       & KaMIS \cite{Lamm2017.Finding}      &                              & 425.96                     & 0.00\%                           & 31.14                               & 45.25               & 0.00\%           & 274.96            & 381.31              & 0.00\%           & 611.79            \\
                                                                     & Gurobi* \cite{gurobi}              &                              & 425.95                     & 0.00\%                           & 3.12                                & 41.38               & 8.55\%           & 23.44             & N/A                 & N/A              & N/A               \\
                \hline
                \multirow{ 2}{*}{\rotatebox[origin=c]{90}{Sup.}}     & Intel* \cite{Li2018.Combinatorial} & TS                           & N/A                        & N/A                              & N/A                                 & 38.80               & 14.25\%          & 9.38              & N/A                 & N/A              & N/A               \\
                                                                     & DGL* \cite{Bother2022.What}        & TS                           & N/A                        & N/A                              & N/A                                 & 37.26               & 17.66\%          & 10.65             & N/A                 & N/A              & N/A               \\
                                                                     & Difusco \cite{Sun2023.DIFUSCOa}    & S                            & 425.30                     & 0.15\%                           & 22.48                               & 40.51               & 10.48\%          & 65.49             & N/A                 & N/A              & N/A               \\
                                                                     & T2T \cite{Li2023.Distribution}     & S                            & \textbf{425.32}                     & \textbf{0.15\%}                           & 24.77                               & \textbf{41.72}               & \textbf{7.80\%}           & 70.74             & N/A                 & N/A              & N/A               \\
                \hline
                \multirow{ 9}{*}{\rotatebox[origin=c]{90}{Not Sup.}} & LwD*    \cite{Ahn2020.Learninga}   & S                            & 422.22                     & 0.88\%                           & 2.26                                & 41.17               & 9.02\%           & 2.97              & 345.88              & 9.29\%           & 28.35             \\
                                                                     & Dimes* \cite{Qiu2022.DIMES}        & S                            & 423.28                     & 0.63\%                           & 2.43                                & 42.06               & 7.05\%           & 5.63              & 332.80              & 12.72\%          & 46.91             \\
                                                                     & GFlowNets**  \cite{Zhang2023.Let}  & S                            & 423.54                     & 0.57\%                           & N/A                                 & N/A                 & 8.53\%           & N/A               & N/A                 & 6.98\%           & N/A               \\
                                                                     & \textsc{Moco}                      & M=1 K=50 b=32                & 422.87                     & 0.73\%                           & 5.76                                & 41.70               & 7.86\%           & 4.19              & 342.69              & 10.13\%          & 134               \\
                                                                     & \textsc{Moco}                      & M=1 K=200 b=32               & 423.27                     & 0.63\%                           & 14.20                               & 42.76               & 5.51\%           & 5.43              & 354.69              & 6.98\%           & 526               \\
                                                                     & \textsc{Moco}                      & M=32 K=50 b=32               & 424.75                     & 0.28\%                           & 68.54                               & 43.69               & 3.45\%           & 7.80              & N/A                 & N/A              & N/A               \\
                                                                     & \textsc{Moco}                      & M=32 K=200 b=32              & 425.02                     & 0.22\%                           & 264.44                              & 44.64               & 1.35\%           & 21.52             & \textbf{363.63}              & \textbf{4.64\%}           & 18,053            \\
                                                                     & \textsc{Moco}                      & M=32 K=50 b=128              & \textbf{425.10}                     & \textbf{0.20\%}                           & 310.97                              & 44.41               & 1.85\%           & 17.85             & N/A                 & N/A              & N/A               \\
                                                                     & \textsc{Moco}                      & M=32 K=200 b=128             & N/A                        & N/A                              & N/A                                 & \textbf{44.87}               & \textbf{0.85\%}           & 58.50             & N/A                 & N/A              & N/A               \\
            \end{tabular}
        }
    \end{threeparttable}
\end{table}
\begin{table}
    \centering
    \caption{TSP results for hybrid heuristics, allowing for local or large neighborhood search. *Result reported from the original paper.}\label{tab:hybrid}
    \resizebox{\textwidth}{!}{
        \begin{tabular}{lll|lllllllll} 
                                                                 & \multirow{ 2}{*}{Method }             & \multirow{ 2}{*}{Dec. Str.} & \multicolumn{3}{c}{100} & \multicolumn{3}{c}{200} & \multicolumn{3}{c}{500}                                                                                                                       \\
                                                                 &                                       &                             & Cost $\downarrow$       & Gap $\downarrow$        & Time $\downarrow$       & Cost $\downarrow$ & Gap $\downarrow$ & Time $\downarrow$ & Cost $\downarrow$ & Gap $\downarrow$ & Time $\downarrow$ \\
            \hline
            \multirow{ 2}{*}{\rotatebox[origin=c]{90}{OR}}       & LKH3    \cite{lkh3}                   &                             & 7.7609                  & 0.000\%                 & 5.19                    & 10.719            & 0.00\%           & 6.37              & 16.555            & 0.00\%           & 10.28             \\
                                                                 & Farthest Insertion                    &                             & 9.9200                  & 27.819\%                & N/A                     & 13.954            & 30.17\%          & N/A               & 21.753            & 31.40\%          & N/A               \\
            \hline
            \multirow{ 4}{*}{\rotatebox[origin=c]{90}{Sup.}}     & Difusco  \cite{Sun2023.DIFUSCOa}      & S + 2-opt                   & 7.7659                  & 0.064\%                 & 1.59                    & N/A               & N/A              & N/A               & 16.681            & 0.76\%           & 7.98              \\
                                                                 & T2T        \cite{Li2023.Distribution} & S + 2-opt                   & 7.7638                  & 0.036\%                 & 1.84                    & N/A               & N/A              & N/A               & 16.621            & 0.40\%           & 4.58              \\
                                                                 & LEHD  \cite{Luo2023.Neural}           & RRC 100                     & 7.7624                  & 0.019\%                 & 17.27                   & 10.726            & 0.06\%           & 34.98             & 16.602            & 0.29\%           & 86.99             \\
                                                                 & LEHD   \cite{Luo2023.Neural}          & RRC 1000                    & \textbf{7.7613}                  & \textbf{0.005\%}                 & 171.39                  & \textbf{10.721}            & \textbf{0.02\%}           & 349.81            & \textbf{16.571}            & \textbf{0.10\%}           & 869.88            \\
            \hline
            \multirow{ 4}{*}{\rotatebox[origin=c]{90}{Not Sup.}} & Dimes   \cite{Qiu2022.DIMES}          & AS + S + MCTS               & 7.7617*                 & 0.010\%*                & N/A                     & N/A               & N/A              & N/A               & 16.827            & 1.64\%           & 94.93             \\
                                                                 & DeepAco  \cite{Ye2023.DeepACOa}       & NLS                         & 7.7632                  & 0.029\%                 & 19.33                   & N/A               & N/A              & N/A               & 16.749            & 1.18\%           & 743.63            \\
                                                                 & \textsc{Moco} + 2-opt                 & M=1 K=200 b=32              & 7.7681                  & 0.092\%                 & 1.28                    & 10.799            & 0.75\%           & 2.70              & 16.757            & 1.22\%           & 16.88             \\
                                                                 & \textsc{Moco} + 2-opt                 & M=32 K=200 b=32             & \textbf{7.7610}                  & \textbf{0.000\%}                 & 13.43                   & \textbf{10.724}            & \textbf{0.04\%}           & 14.90             & \textbf{16.621}            & \textbf{0.40\%}           & 276.40            \\
        \end{tabular}%
    }
\end{table}%
In \cref{tab:hybrid}, we compare to hybrid approaches that combine learning-based methods with traditional heuristics such as local search, large neighborhood search or decomposition. For \textsc{Moco} we consider a simple hybrid method, that runs a 2-opt after construction, similar to \cite{Ye2023.DeepACOa}. Note that we use a naive CPU implementation and runtime can be improved by also parallelizing this operation on the GPU.
\section{Conclusion}
We have proposed \textsc{Moco}, a learnable optimizer for NP-hard binary graph
combinatorial problems. \textsc{Moco} uses two
GNNs. The first initializes a heatmap $\theta$. Then, iteratively, solutions
are constructed from $\theta$ and a second GNN updates $\theta$ based on
meta-features extracted from the solutions. \textsc{Moco} outperforms or is competitive with other learning based approaches. We showcase additional favorable properties such as learning different exploration-exploitation trade-offs for different training budgets and generalization behavior to configurations not seen during training enabling inference on larger budgets and batch sizes.
\begin{credits}
    \subsubsection{\ackname} This work was supported by the state of Lower Saxony under the project "Digitale Lehre Hub Niedersachsen - KI in Studium, Lehre und Prüfungen".
\end{credits}


\bibliographystyle{splncs04}
\bibliography{MoCo}

\newpage
\appendix

\section{Traveling Salesman Problem}\label{app:tsp}

A TSP instance consists of a fully connected  graph $G = (V, E)$, where
$V = \{v_1, ..., v_n\}$ is the set of $n$ customers and
$E=\{e_1,\dots,e_{n^2}\}$ is the set of edges. Each edge $e \in E$ is associated
with a distance of $d(e)$ representing the cost of traveling from its starting
point to its end point.
The set of feasible solutions $\Omega_G \subseteq \{0,1\}^{n^2}$  is given by the
vectors $x$ for which the corresponding path in $G$ is a Hamiltonian cycle.
Here, $x_i \in \{0,1\}$ displays whether $e_i \in E$ is in the cycle. The
objective function is given via
\begin{equation*}
    f_G(x) \coloneqq \sum_{i=1}^{n^{2}} x_id(e_i).
\end{equation*}
\section{Maximum Independent Set}\label{app:mis}
Let $G=(V,E)$ be a graph with $n$ nodes. The set of feasible solutions $\Omega_G \subseteq
    \{0,1\}^{n}$ is given by the vectors $x$ for which the corresponding set of vertices
only contains of pairwise not adjacent vertices. Here, $x_i \in \{0,1\}$ displays
whether $v_i$ is in the chosen independent set. The goal is to find such a
vertex set of maximal cardinality. Hence, the objective function is given via

\begin{equation*}
    f_G(x) = - \sum_{i=1}^{n}x_i.
\end{equation*}

\section{Ablation Study}\label{app:ablation}
Additionally, we show an ablation experiment on the TSP200 in Table \ref*{tab:ablation}, where we test replacing the first GNN with a simple heuristic: $$\theta^{(0)}_{ij}=-d_{ij}$$ where $d_{ij}$ is the distance between node $i$ and $j$ making it less likely to go from node $i$ to $j$ in the tour, the larger the distance between the nodes is. Then we update $\theta$ by various mechanisms, showing that our proposed components are the contributing factors of the performance. Firstly, we update directly with Adam. Secondly we update with a learned MLP optimizer \cite{Metz2019.Understanding} which only takes into account the parameters, gradients, momentum and generic global features such as the budget and step, but neglects the graph structure of the problem and related additional features, acting only independently per parameter. Thirdly, we add our learnable GNN based optimizer but with the heuristic initialization scheme and finally we evaluate our full method. Hyperparameters are optimized for all methods. The results are computed over an additionally generated evaluation set and the relative gap to LKH3 is reported.  

\begin{table}
    \centering
    \caption{Ablation study on the TSP200.}\label{tab:ablation}
        \begin{tabular}{ll|lllll}
            Initialization & Update & $M$   & $K$    & $b$   & Cost  & Gap   \\
            \hline
            heuristic      & Adam   & 1   & 200  & 32  & 29.13   & 174\% \\
            heuristic      & Adam   & 1   & 1000 & 128 & 19.08   & 79.5\%  \\
            heuristic      & MLP    & 1   & 200  & 32  & 11.90  & 11.9\%  \\
            heuristic      & GNN    & 1   & 200  & 32  & 10.86 & 2.22\%   \\
            GNN            & GNN    & 1   & 200  & 32  & 10.84 & 2.05\%   \\
        \end{tabular}
\end{table}

\section{Meta Optimizer Architecture}\label{app:architecture}
Our meta optimizer consists of two GNNs, one for initializing $\theta$ and one for updating $\theta$ in between rollouts. Both GNNs share the same architecture and operate on the same graph structure, stemming from the instances' graph $G$. The features differ, as the initializing GNN operates only on the features that describe the problem instance while for updating $\theta$, the additional features described in section \ref*{sec:meta_optimizer} are added. In the following we describe the architecture used for the TSP and MIS.
\subsection{Traveling Salesman Problem}
Next to the generic features described in section \ref*{sec:meta_optimizer}, we add two additional features in order to describe the instance.
\begin{enumerate}
    \item The distance $d_{ij} \in \mathbb{R}$ for each edge $e_{ij}$.
    \item A binary node feature $s_i \in \{0,1\}$ indicating whether node $i$ represents the starting node of the construction process. We sample the starting position randomly but then keep it fixed for the whole optimization procedure over $\theta$.
\end{enumerate}
For the GNN architecture, we adopt the \textit{GraphNetwork} framework from \cite{Battaglia2018.Relational}. 
Let $h^v_i, h^e_{ij}, h^g \in \mathbb{R}^{d_{\text{hidden}}}$ denote the node features, edge features and optionally global features, after an initial linear embedding layer to the embedding dimension $d_{\text{hidden}} = 128$.

A \textit{GraphNetwork} block then sequentially updates first the edge, then the node and finally the global embeddings. The edge update function is given by:
\begin{equation}
    \Tilde{h}^{e}_{ij} = h^e_{ij} + \text{LN}(\text{RELU}(W_e[h^e_{ij}; h^v_i; h^v_j; h^g] + b_e))
\end{equation}
where $W_e \in \mathbb{R}^{d_{\text{hidden}} \times 4d_{\text{hidden}}}$ and $b_e \in \mathbb{R}^{d_{\text{hidden}}} $ are learnable weights, $;$ is the concatenation operator and $\text{LN}$ is a layer normalization per feature over all edge embeddings. The node update function is given by:
\begin{equation}
    \Tilde{h}^{v}_{i} = h^v_{i} + \text{LN}(\text{RELU}(W_v[h^v_i; \Bar{h}^{e}_{s_i}; \Bar{h}^{e}_{r_i}; h^g] + b_v))
\end{equation}
where $W_v \in \mathbb{R}^{d_{\text{hidden}} \times 4d_{\text{hidden}}}$ and $b_v \in \mathbb{R}^{d_{\text{hidden}}} $ are learnable weights, $\Bar{h}^{e}_{s_i}$ and $\Bar{h}^{e}_{r_i}$ are the aggregated updated edge embeddings of all outgoing and incoming edges from node $i$ respectively:
\begin{equation}
    \Bar{h}^{e}_{s_i} = \sum_{j}{\Tilde{h}^{e}_{ij}} \quad \text{and} \quad \Bar{h}^{e}_{r_i} = \sum_{j}{\Tilde{h}^{e}_{ji}}
\end{equation}
Aggregation is done via an element-wise summation. Finally, the global update function is given by:
\begin{equation}
    \Tilde{h}^{g} = h^g + \text{RELU}(W_g[\Bar{h}^{v}; \Bar{h}^{e}; h^g] + b_g)
\end{equation}
where $W_g \in \mathbb{R}^{d_{\text{hidden}} \times 3d_{\text{hidden}}}$ and $b_g \in \mathbb{R}^{d_{\text{hidden}}} $ are learnable weights and $\Bar{h}^{v}$ and $\Bar{h}^{e}$ are the aggregated node and edge embeddings of the entire graph:
\begin{equation}
    \Bar{h}^{v} = \sum_{i \in V}{\Tilde{h}^{v}_{i}} \quad \text{and} \quad \Bar{h}^{e} = \sum_{(i,j) \in E}{\Tilde{h}^{e}_{ij}}
\end{equation}
We stack $L=3$ of these blocks and then apply a final linear layer to decode the edges for the heatmap $\theta$, as described in section \ref*{sec:meta_optimizer}. Since the graph for initializing $\theta$ does not have global features, the global update is omitted and the global features are removed from node and edge update functions.

\subsection{Maximum Independent Set}
Since the MIS graph does not add any additional features, we add dummy node features to the graph with constant value 1. As the decision variables in the MIS problem lie on the nodes of the graph, and there are no edge features, we employ a Graph Convolutional Network. We use the GCN architecture \cite{Kipf2017.Semisupervised} with slight modifications. We use a different weight matrix for self connections and add a residual connection leading to the node update function:
\begin{equation}
    \hat{h}^{v}_{i} = h^v_i + \text{RELU}(W_1 h^v_i + b_1 + \max_{j \in \mathcal{N}(i)}{(W_2 h^v_j + b_2)})
\end{equation}
where $W_1, W_2 \in \mathbb{R}^{d_{\text{hidden}} \times d_{\text{hidden}}}$ and $b_1, b_2 \in \mathbb{R}^{d_{\text{hidden}}} $ are learnable weights. The $\max$ operator is an element-wise aggregator of the neighborhood.
If there are global features, we further update the nodes
\begin{equation}
    \tilde{h}^{v}_{i} = \hat{h}^{v}_{i} + \text{RELU}(W_3 [h^g;\hat{h}^{v}_{i}] + b_3)
\end{equation}
as well as the global features themselves
\begin{equation}
    \tilde{h}^{g} = h^g + \text{RELU}(W_4 \sum_{i \in V}{\Tilde{h}^{v}_{i}} + b_4)
\end{equation}
We stack $L=3$ of these blocks and then apply a final linear layer to decode the nodes for the heatmap $\theta$, as described in section \ref*{sec:meta_optimizer}.

\section{Meta Training Details}\label{app:meta_training}
To estimate the descent direction of the GNN parameters, we use evolutionary strategies as described in \cref*{sec:meta_training}. In order to do so, we draw $N=128$ perturbations for TSP and $N=64$ perturbations for larger for MIS. For updating the GNN parameters, the estimated descent directions from ES are processed by Adam. We use a learning rate of 0.001 with a cosine learning rate decay schedule and 50 linear warm up steps to the maximum learning rate. From there the learning rate is fully annealed to 0 at the end of training in one cycle. For the larger TSP variants we find it beneficial to optimize over the logarithm of the meta loss, in order to avoid large differences in gradient magnitudes that stem from the large objective values of poor random TSP solutions. Alternatively, this problem could be addressed by clipping the loss difference between the antithetic pairs in equation \cref*{eq:es_estimator} to a maximum absolute value.

Our method is implemented in Jax \cite{jax2018github}. For the TSP, all models were trained for $K=200$ and $b=32$. Likely, more efficient training strategies could be employed, such as a training schedule for $K$ and warm-starting from smaller sizes. We leave this for future work. For the MIS, on the ER[700-800] graphs, we first train a model with $K=50$ and $b=32$ and then a second model with $K=200$ and $b=32$ was finetuned from the first model. For the ER[9000-11000] graphs, we rely on the models trained on the smaller graphs. For SATLIB, we only train one model with $K=50$ and $b=32$.

\section{Baselines}\label{app:baselines}
We provide brief descriptions of the baselines used in our experiments.
\subsection*{Dimes \cite{Qiu2022.DIMES}}
Dimes uses a GNN to initialize a heatmap $\theta$ and then uses a policy gradient method to further optimize the MLP decoder layer of the GNN that creates the heatmap, given a single instance. Optionally, the MCTS k-opt local search from \cite{DBLP:conf/aaai/FuQZ21} 
can be used to further improve the solution. The model is trained with Maml based meta REINFORCE gradients.

Note that for the MIS problem Dimes only samples from the initial heatmap without the gradient based improvement steps, stating that they could not improve the solutions on this problem through the Active Search component. In our experiments, with the additional features and learned update, we find it helpful to do multiple update steps.
\subsection*{DeepACO/ACO \cite{Ye2023.DeepACOa}}
DeepACO builds on top of traditional AntColony Optimization (ACO). ACO uses $b$ \emph{ants} to construct solutions from a vector of \emph{pheromones} and \emph{heuristic components} over the decision variables. \emph{Heuristic components} are fixed for a problem instance and \emph{pheromones} are updated by the ants after construction. Together they can be considered making up what we have called $\theta$. DeepACO uses a GNN to learn the \emph{heuristic components} but leave the update procedure via \emph{pheromones} unchanged. The model is trained with a combination of RL and imitation learning. Optionally, a 2-opt local search can be used to further improve the solution. Their NLS variant uses local search to iteratively change between performing local search on actual cost matrix and the learned \emph{heuristic components}.
\subsection*{Difusco \cite{Sun2023.DIFUSCOa}}
Difusco uses a GNN to learn a heatmap, modeling the probability of selecting each decision variable independently. They decode this heatmap greedily by sorting the heatmap values by maximal probability and deciding on the variables with the highest probability, if there are no conflicts with the constraints. The model is trained with supervised learning via diffusion from a large dataset of high quality solutions. Optionally, a 2-opt local search can be used to further improve the solution. The diffusion model samples multiple heatmaps at inference time to search through the solution space.
\subsection*{T2T \cite{Li2023.Distribution}}
T2T builds on Difusco, adding a search procedure based on noising a heatmap constructed from the best solution found from the first decoding. The subsequent denoising is guided by an additional approximate surrogate loss that is constructed per problem type.
\subsection*{Pomo \cite{Kwon2020.POMO}}
Pomo uses the encoder-decoder architecture of \cite{DBLP:conf/iclr/KoolHW19} to construct solutions for COPs. The model is trained with RL, and they improve the baseline computation by making use of the solution symmetricity in the TSP, since all solutions are equivalent up to the starting point. They also provide an inference mechanism of decoding the model greedily for each node in the problem as a different starting node and for 8 different rotations of the coordinates.
\subsection*{Pointerformer \cite{Jin2023.Pointerformer}}
The Pointerformer modifies the encoder-decoder architecture of \cite{DBLP:conf/iclr/KoolHW19} with a reversible encoder to avoid OOM problems with long unrolls. Additionally, they modify the context vector and logit clipping of the decoder. They utilize the same training and inference procedure as Pomo \cite{Kwon2020.POMO} and train their model on up to 500 nodes.
\subsection*{EAS \cite{DBLP:conf/iclr/HottungKT22}}
Efficient Active Search (EAS) proposes to update a subset of the model parameters during search. They use policy gradients from sampled solutions, as well as the cross-entropy loss from the current best solution.
\subsection*{SGBS \cite{DBLP:conf/nips/ChooKKJHTG22}}
Simulation-Guided Beam Search (SGBS) proposes a tree search algorithm for combinatorial optimization similar to beam search based on fully rolling out a subset of promising candidates and then reevaluating the candidates based on the achieved scores. The method is further combined with EAS \cite{DBLP:conf/iclr/HottungKT22}.
\subsection*{BQ  \cite{Drakulic2023.BQNCO}}
BQ-Transformer replaces the encoder-decoder model by a transformer model, that recomputes the embedding at every construction step. They show that this leads to much better size extrapolation performance. A drawback however is, that the high computational requirements of the model, leads them to train it with supervised learning from solver obtained solutions even for smaller problems such as TSP instances with 100 nodes.
\subsection*{LEHD \cite{Luo2023.Neural}}
LEHD uses a similar architecture to \cite{Drakulic2023.BQNCO}, also relying on supervised learning. Additionally, they combine their model with a Large Neighborhood Search strategy that iteratively reoptimizes subsegments of the current solution, called Random Reconstruct (RRC).
\subsection*{Intel \cite{Li2018.Combinatorial}}
They learn a GNN to predict $m$ heatmaps, each assigning a probability per node of that node belonging to the maximal independent set. The model is trained with supervised learning, via classification from optimal solutions. For inference, they use the heatmaps to guide a tree search algorithm. The search inserts nodes, according to maximal probability, until a conflict would arise, at which point, the resulting partial solution is added to a queue and the heatmap is recomputed for the remaining graph. The search continues until the queue is empty, and at every step all $m$ heatmaps are expanded. Optionally, MIS specific graph reduction techniques and additional local search can be used.
\subsection*{LwD \cite{Ahn2020.Learninga}}
LwD learns a GraphSage agent via Proximal Policy Optimization to solve MIS problems. Instead of iteratively only deciding on a single node, they reformulate the MDP, such that the agent at each step can decide for multiple nodes that they belong or do not belong to the independent set or to defer them to the next steps. Arising conflicts are resolved by the environment through an update and clean up phase.

\end{document}